\documentclass[10pt,twocolumn,letterpaper]{article}

\usepackage{amsmath}
\usepackage{amssymb}
\usepackage{mathtools}
\usepackage{amsthm}

\usepackage{multirow}

\theoremstyle{plain}
\newtheorem{theorem}{Theorem}[section]
\newtheorem{proposition}[theorem]{Proposition}

\theoremstyle{definition}
\newtheorem{definition}[theorem]{Definition}

\theoremstyle{remark}
\newtheorem{remark}[theorem]{Remark}

\newcommand\R{\mathbb{R}}

\usepackage[pagenumbers]{iccv} 

\definecolor{iccvblue}{rgb}{0.21,0.49,0.74}
\usepackage[pagebackref,breaklinks,colorlinks,allcolors=iccvblue]{hyperref}

\def\paperID{12219} 
\def\confName{ICCV}
\def\confYear{2025}

\title{Enhancing Transformers Through Conditioned Embedded Tokens}

\author{
\begin{tabular}{c@{\hskip 1in}c}
Hemanth Saratchandran & Simon Lucey \\
\small\texttt{hemanth.saratchandran@adelaide.edu.au} & \small\texttt{simon.lucey@adelaide.edu.au}
\end{tabular}
\\[1.5ex]
Australian Institute for Machine Learning, University of Adelaide
}

\begin{document}

\twocolumn[{%
\renewcommand\twocolumn[1][]{#1}%
\maketitle
\begin{center}
    \centering
    \captionsetup{type=figure}
    \includegraphics[width=1\linewidth]{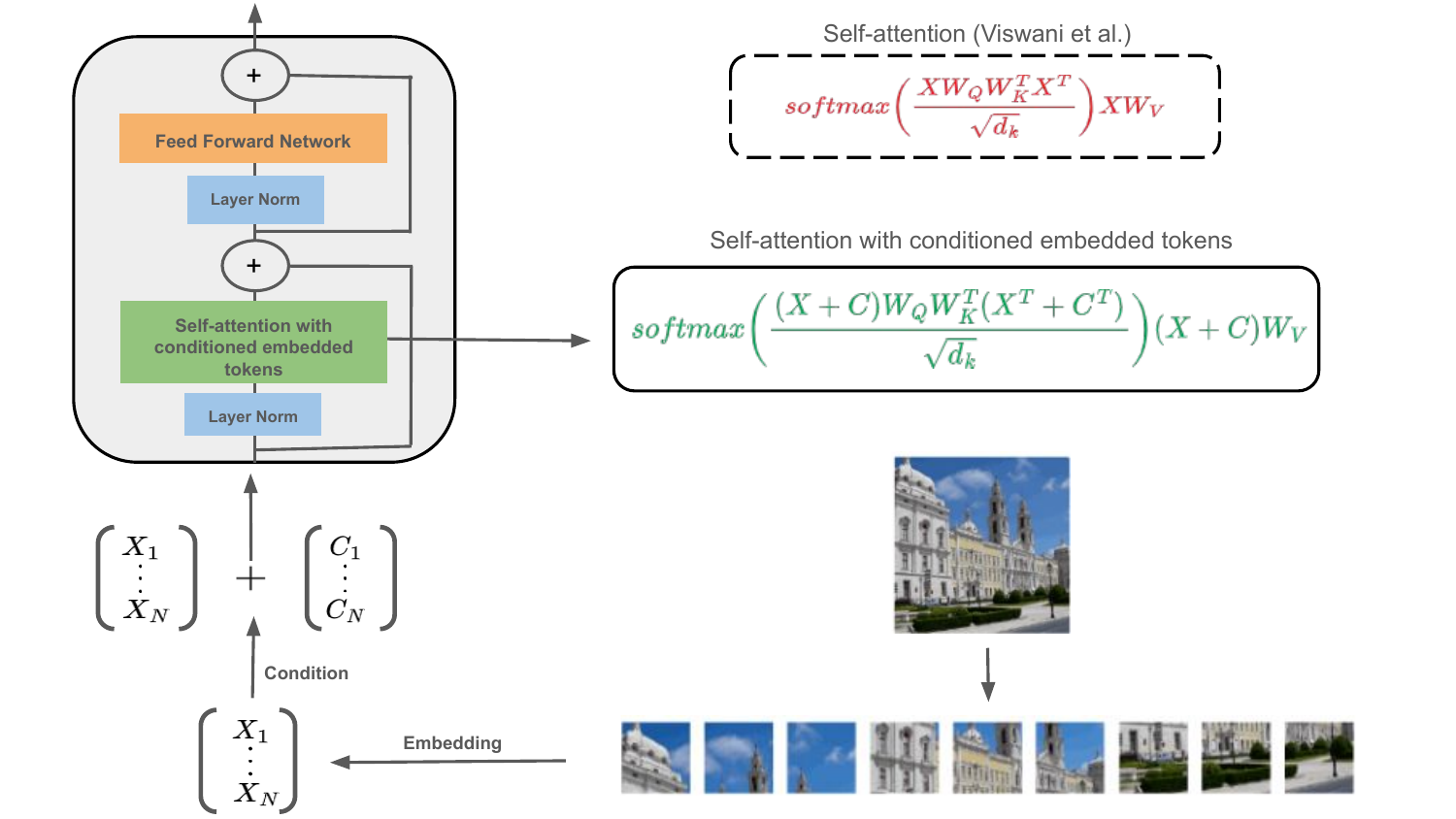}
\captionof{figure}{\textbf{Schematic representation of conditioned embeddings for a vision transformer:} An image is divided into \( N \) patches, with each patch then embedded as a high-dimensional vector \( X_i \in \mathbb{R}^d \) for \( 1 \leq i \leq N \). These vectors are then concatenated to form the embedded token matrix \( X = [X_1 \cdots X_N]^T \). To improve the conditioning of the self-attention mechanism, a correction term \( C = [C_1 \cdots C_N]^T \) is added to \( X \), reducing its condition number. The modified matrix is then fed into the first layer of the transformer (positional encoding not shown). Its effect on the self-attention equation is illustrated in the green equation (for simplicity, layer normalization has been omitted).}
    \label{fig:front_fig}
\end{center}%
}]

\begin{abstract}
Transformers have transformed modern machine learning, driving breakthroughs in computer vision, natural language processing, and robotics. At the core of their success lies the attention mechanism, which enables the modeling of global dependencies among input tokens. However, we reveal that the attention block in transformers suffers from inherent ill-conditioning, which hampers gradient-based optimization and leads to inefficient training. To address this, we develop a theoretical framework that establishes a direct relationship between the conditioning of the attention block and that of the embedded tokenized data. Building on this insight, we introduce conditioned embedded tokens, a method that systematically modifies the embedded tokens to improve the conditioning of the attention mechanism. Our analysis demonstrates that this approach significantly mitigates ill-conditioning, leading to more stable and efficient training. We validate our methodology across various transformer architectures, achieving consistent improvements in image classification, object detection, instance segmentation, and natural language processing, highlighting its broad applicability and effectiveness.
\end{abstract}    
\section{Introduction}
\label{sec:intro}

The transformer architecture \cite{vaswani2017attention} has become a cornerstone of modern machine learning, driving breakthroughs across various domains including computer vision \cite{dosovitskiy2020image, liu2021swin, touvron2021training, carion2020end}, natural language processing \cite{vaswani2017attention, zhuang2021robustly, zhen2022cosformer}, and robotics \cite{salzmann2020trajectron++, maiti2023transfusion}. Its success is largely attributed to its ability to model complex relationships in data without relying on traditional recurrent or convolutional structures. At the heart of the transformer is the self-attention mechanism, which dynamically assigns importance to different parts of the input, capturing both local and global dependencies. Unlike traditional architectures that process inputs in a fixed order, transformers utilize self-attention to integrate global context at each layer, making them highly effective for tasks that require contextual understanding.

In general, transformers process input data by first converting it into discrete tokens, which are then mapped to a high-dimensional vector space through an embedding layer. For text, these tokens represent words or sub-words, while for images, image patches are treated as tokens. The resulting embedded tokens form a matrix that is passed into the first transformer layer for self-attention computations. This process allows the model to effectively learn relationships between tokens in a high-dimensional space.

In this paper, we explore the conditioning of the self-attention matrix in transformer architectures—an essential yet often overlooked factor influencing optimization dynamics. The conditioning of a matrix is measured by its condition number, defined as the ratio of its largest singular value to its smallest. A high condition number indicates ill-conditioning, which is a well-known challenge for the convergence of gradient-based optimization methods \cite{nocedal1999numerical}.

Previous work has addressed conditioning in feedforward neural networks. In \cite{saratchandran2025weight}, a methodology was introduced to condition network weights, leading to improved optimization. Similarly, \cite{liu2022loss} examined the condition of the neural tangent kernel (NTK), proposing a framework for reducing its condition number and demonstrating enhanced convergence for gradient descent. While these studies have focused on conditioning in feedforward architectures, its role in self-attention mechanisms remains largely unexplored. This gap motivates our investigation into how conditioning affects optimization in transformers.


We develop a theoretical framework to analyze the condition number of the self-attention matrix within a transformer's attention layer. Our analysis establishes a direct relationship between the condition number of the first-layer self-attention matrix and that of the embedded tokens. Leveraging this insight, we introduce conditioned embedded tokens, a method that applies a structured correction term to the embedded token matrix, effectively reducing its condition number and, in turn, improving the conditioning of the self-attention matrix. While a full theoretical proof linking this improvement to better convergence in gradient-based optimization would require analyzing the NTK of a transformer—a notoriously difficult task—we provide strong empirical evidence that conditioned embedded tokens enhances performance across a range of transformer applications.


\Cref{fig:front_fig} illustrates how conditioned embedded tokens are integrated into a vision transformer. An image is first divided into tokens via patches, which are vectorized and passed through a learnable embedding layer, producing embedded tokens in a high-dimensional vector space. These tokens form the matrix \( X = [X_1 \cdots X_N]^T \). A correction term \( C = [C_1 \cdots C_N]^T \) is added to \( X \), reducing its condition number before the modified embedding matrix $X + C$ is fed into the first transformer layer (positional encoding not shown).  
This correction not only lowers the condition number of the self-attention matrix in the first layer but also has a cascading effect, improving conditioning in subsequent layers. In general, positional encodings are added to the tokens before they are fed into the first transformer layer; however, they are omitted in \Cref{fig:front_fig} for simplicity.
Our theory will show that the 
optimal correction term can be derived from the singular value decomposition (SVD) of $X$. 

We validate our framework across various applications, including image classification, object detection, instance segmentation, and natural language processing. A key advantage of conditioned embedded tokens is their flexibility—they can be used with a wide range of attention mechanisms, making them compatible with modern transformer architectures. As we demonstrate, our methodology can serve as a drop-in enhancement for advanced attention mechanisms proposed in recent works \cite{ali2021xcit, liu2021swin, touvron2021training, yuan2022volo, ding2022davit, xiong2021nystromformer}, offering a simple yet effective improvement to existing models. Our main contributions include:
\begin{itemize} 
\item[1.] A theoretical framework introducing conditioned embedded tokens for improving the conditioning of the self-attention matrix within the first transformer layer yielding a more stable transformer architecture.
\item[2.] An empirical evaluation of conditioned embedded tokens across various transformer architectures and applications, including image classification, object detection, instance segmentation, and natural language processing, showing in each case its superior performance.
\end{itemize}

\section{Related Work}\label{sec:rel_work}

\paragraph{Attention:} Various strategies have been introduced to enhance the efficiency and effectiveness of transformers, particularly by addressing their computational overhead and rethinking attention mechanisms. The Data-Efficient Image Transformer (DeiT) \cite{touvron2021training} improves training efficiency in vision transformers by leveraging distillation tokens, allowing for competitive performance without requiring massive datasets. The Cross-Covariance Image Transformer (XCiT) \cite{ali2021xcit} redefines attention by operating on cross-covariances of spatial features, enabling more efficient spatial interactions while reducing computational demands. 
Another notable approach, the Nyströmformer \cite{xiong2021nystromformer}, approximates standard self-attention using the Nyström method, reducing quadratic complexity to near-linear while preserving key attention properties. This technique has proven particularly effective for long-range sequence modeling.
In this paper, we demonstrate that our conditioning framework can be integrated as a drop-in replacement into these advanced attention mechanisms, yielding superior performance across multiple architectures.

\textbf{Conditioning:} In \cite{saratchandran2025weight}, the authors investigate the condition number of weight matrices in feedforward neural networks and demonstrate that lower condition numbers lead to improved accuracy across various applications. They propose a preconditioner matrix that multiplies each layer’s weight matrix, effectively reducing its condition number and enhancing the training of dense weights.
In \cite{liu2022loss}, neural network optimization is explored through the lens of the neural tangent kernel. The authors introduce the concept of a PL* condition, a variant of the Polyak-Lojasiewicz condition, and show that this approach improves the conditioning of the neural tangent kernel, which, in the infinite-width limit, governs the dynamics of gradient descent \cite{jacot2018neural}.
For feedforward networks, \citet{agarwal2021deep} showed that increasing depth improves conditioning and thereby facilitates optimization, while \citet{macdonald2023skip} demonstrated that normalization can further enhance conditioning. In the context of implicit neural representations (INRs), several works have established that activation choice strongly influences conditioning and optimization performance \citep{saratchandran2023curvature, chng2022gaussian, saragadam2023wire, gao2024h, saratchandran2024activation, chng2024invertible, ramasinghe2022you, saratchandran2024sampling}. For transformers, skip connections have been shown to improve the conditioning of the attention block \citep{ji2025always}, while multi-head attention and polynomial activations have been shown to condition the Jacobian of the attention mechanism more effectively \citep{saratchandran2025leaner, saratchandran2024rethinking}. Finally, in fine-tuning settings, conditioning adapter modules has been found to yield faster and more stable convergence \citep{ji2024efficient, albert2025randlora, albert2025towards}.

\section{Notation}\label{sec:notation}
In this section, we formally define the transformer architecture by describing the structure of a transformer layer, along with establishing notation for various mathematical elements that will be referenced in subsequent sections. For further foundational details on transformers, readers may consult \citet{vaswani2017attention}.

A layer in a transformer can be represented as a mapping 
\begin{equation}
  \mathbf{T}: \mathbb{R}^{N \times D} \rightarrow \mathbb{R}^{N \times D}  
\end{equation}
defined by 
\begin{equation}\label{eqn:trans_main}
    \mathbf{T}(X) = \mathbf{F}(\mathbf{A}(X) + X),
\end{equation}
where \( \mathbf{F} \) denotes a feed forward network with a residual connection, and \( \mathbf{A} \) represents an attention mechanism. In general, layer normalization is added however for simplicity we omit layer normalization for this discussion.

To begin with input data $x$ is tokenized into $N$ tokens yielding a tokenized representation $(x_1, \cdots ,x_N)$ where $x_i \in \R^{t \times 1}$. Each $x_i$ is then mapped to a high dimensional space $\R^{d \times 1}$ using a learnable embedding layer $E \in \R^{d \times t}$ so that 
\begin{equation}
Ex_i \in \R^{d \times 1}.    
\end{equation}
The matrix $[Ex_1\cdots Ex_n]^T \in \R^{N\times d}$ represents the embedded tokens associated to the input data. To capture the order of the tokens, a positional encoding matrix $P \in \R^{N\times d}$ is added forming $X = [Ex_1\cdots Ex_n]^T + P \in \R^{N \times d}$. The matrix $X$ is then input into the first layer of the transformer. 
For the theoretical \cref{sec:theory} we will often use the notation $X$ to denote embedded tokens often leaving out explicit mention of the positional encoding part for ease of notation.  


The key component of the transformer is self-attention. This is  composed of three learnable matrices, a query (\( W_Q \)), key (\( W_K \)), and value (\( W_V \)), where \( W_Q, W_K \in \mathbb{R}^{D \times d} \) and \( W_V \in \mathbb{R}^{D \times d} \). The output of the attention head \( \mathbf{A}(X) \) is then given by
\begin{equation}\label{eqn:attn_eqn_general}
    \mathbf{A}(X) = \mathbf{softmax}(XW_QW_k^TX^T)XW_V,
\end{equation}
where $\mathbf{softmax}$ is the softmax activation that acts row-wise on the matrix $XW_QW_k^TX^T$. Note that then $A(X) \in \R^{N\times d}$. In the case no activation is used we have the definition of linear self-attention which we will denote by
\begin{equation}
\mathbf{LA}(X) := XW_QW_k^TX^TXW_V.
\end{equation}

In general, multiple attention heads \( \mathbf{A}_i \) for \( 1 \leq i \leq h \) are utilized, each of dimension $N \times d_i 
= N \times \frac{d}{h}$.
These are then concatenated to produce a multi-head attention output,
\begin{equation}\label{eqn:multi_head_eqn}
[\mathbf{A}_1, \cdots, \mathbf{A}_h],
\end{equation}
which is subsequently fed into a feedforward layer. Similary, a multi-head linear attention can be defined in the same way. The full transformer architecture is obtained by sequentially stacking several such transformer layers, as defined in \cref{eqn:trans_main}.

\section{Theoretical Framework}\label{sec:theory}

\subsection{Conditioning of self-attention}

In this section, we analyze the conditioning of the self-attention matrix in a transformer. Our objective is to demonstrate that the condition number of the self-attention matrix in the first layer is influenced by the condition number of the embedded tokens. This insight motivates our key finding: reducing the condition number of the embedded tokens leads to a better-conditioned self-attention matrix in the first layer, ultimately improving the overall conditioning of the transformer. In this section, we omit positional encoding and layer normalization to keep the core theoretical insight clear and accessible. In \cref{sec:exps}, we will show our insights go through for a variety of practical transformers that use positional encoding and layer normalization.

\begin{definition}
Let $A$ be an $N \times d$ matrix of full rank. The condition number of $A$, denoted by $\kappa$, is defined as
\begin{equation}
    \kappa(A) = \frac{\sigma_1}{\sigma_k}
\end{equation}
where $k = \min\{N, d\}$, and $\sigma_1$ denotes the largest singular value of $A$ and $\sigma_k$ the smallest singular value of $A$. Note that as $A$ is assumed full rank $\sigma_k \neq 0$ and the condition number is well-defined. Furthermore, observe that $\kappa(A) \geq 1$ since 
$\sigma_1 \geq \sigma_k$. 
\end{definition}

We have the following estimate for the condition number of linear self-attention and softmax self-attention.

\begin{proposition}\label{prop:attention_cond_upper}
Let $X$ denote an input for an attention layer as defined in \cref{sec:notation}. Assume $\mathbf{LA}(X)$ and $\mathbf{A}(X)$ have full rank. We then have    
\begin{align}
 \kappa(\mathbf{LA}(X)) &\leq \kappa(W_Q)\kappa(W_K)\kappa(W_V)\kappa(X)^3 \label{eqn:linear_cond_bound}\\
\kappa(\mathbf{A}(X)) &\leq \kappa(\mathbf{softmax}(XW_QW_K^TX^T))\kappa(X)\kappa(W_V) \label{eqn:sm_cond_bound}.
\end{align}
\end{proposition}
The proof of Proposition \ref{prop:attention_cond_upper} is given in  Sec. A of the Supp. material.


\begin{remark}\label{rmk:rank_assump}
The statement of Proposition \ref{prop:attention_cond_upper} assumes the attention matrices are full rank. Since the layers of a transformer are randomly initialized we have that with probability $1$ the attention matrices will be full rank at initialization.
\end{remark}


We point out that although \cref{eqn:linear_cond_bound} and \cref{eqn:sm_cond_bound} only provide an upper bound for the condition number, we found that the this upper bound provides a useful measure for estimating the condition number of the self-attention matrix. Therefore, for the self-attention matrices $\mathbf{A}(X)$ and $\mathbf{LA}(X)$  we make the following definition
\begin{align}
 \mu(\mathbf{LA}(X)) &:= \kappa(W_Q)\kappa(W_K)\kappa(W_V)\kappa(X)^3. \label{eqn:lin_att_complex}\\  
 \mu(\mathbf{A}(X)) &:= \kappa(\mathbf{softmax}(XW_QW_k^TX^T))\kappa(X)\kappa(W_V). \label{eqn:sm_complex}   
\end{align}
 


\subsection{Conditioned embedded tokens}\label{subsec:cond_emb_token}

In this section, we present our main theorem, which provides a method for reducing the condition number of embedded tokens derived from a dataset. This reduction subsequently decreases the measure $\mu(\mathbf{A}(X))$, a key metric for assessing the condition number of the self-attention matrix in the first transformer layer.

\begin{theorem}\label{thm:conditioned_tokens}
Let  $(x_1,\cdots ,x_N)$ denote $N$ tokens associated to an input dataset with $x_i \in \R^{t \times 1}$. Let $E \in \R^{d \times t}$ denote an embedding matrix so that $Ex_i \in \R^{d \times 1}$. Let $X = [Ex_1 \cdots Ex_N]^T \in \R^{N \times d}$ denote the matrix of embedded tokens. 
Assume that $\kappa(X) > 2$.
Then there exists a matrix 
$C \in \R^{N \times d}$ such that
\begin{equation}
     \kappa(X + C) \leq 2
\end{equation}
and hence that $\kappa(X + C) < \kappa(X)$.
\end{theorem}
In particular, \cref{thm:conditioned_tokens} shows that if the condition number of the embedded tokens $X$ is extremely large. Adding a correction term given by $C$ will bring it down significantly. As we shall see in the experiments \cref{sec:exps} the condition number $\kappa(X)$ is empirically extremely large.
The proof of \cref{thm:conditioned_tokens} is given in  Sec. A of the Supp. material. 


\begin{theorem}\label{thm:complexity_lower}
Let $X$ denote a collection of embedded tokens, as defined in the statement of \cref{thm:conditioned_tokens}, and let $C$ denote the matrix given by \cref{thm:conditioned_tokens}. We then have
\begin{equation}
 \mu(\mathbf{LA}(X+C)) \leq \mu(\mathbf{LA}(X)).  
\end{equation}
Assume that 
\begin{align}
 &\kappa(\mathbf{softmax}((X+C)W_QW_K^T(X^T+C^T))) \nonumber\\
 &\leq  
 \kappa(\mathbf{softmax}(XW_QW_K^TX^T))
\end{align}
then 
\begin{equation}
\mu(\mathbf{A}(X+C)) \leq \mu(\mathbf{A}(X))
\end{equation}
where $\mu(\mathbf{LA}(X))$ is defined by \cref{eqn:lin_att_complex} and $\mu(\mathbf{A}(X))$ by \cref{eqn:sm_complex}.
\end{theorem}
The proof of \cref{thm:complexity_lower} is given in Sec. A of the Supp. material. A natural question arises: can self-attention be conditioned directly through queries, keys, and values, independent of the embedded tokens? Unfortunately, our approach of introducing a correction term does not extend to these components. This is because our correction modifies the singular values of the embedded tokens, whereas altering the singular values of queries, keys, or values could disrupt how self-attention captures spatial relationships between tokens.


\begin{table*}[!ht]
    \centering
    \begin{tabular}{c|c|c|c|c|c}

        \midrule
        \multirow{2}{*}{} & \multicolumn{4}{c}{Models} \\
        & ViT-base & DeiT-base & Swin-base & XciT-medium & DaViT-base\\
        \midrule
        Original embedded tokens & 80.3 & 81.6 & 83.1 & 82.2 & 83.6 \\
        \midrule
       Conditioned embedded tokens & \textbf{81.3} & \textbf{82.5} & \textbf{83.9} & \textbf{82.9} & \textbf{84.6} \\
        \midrule
     \end{tabular}
\caption{Comparison of vision transformers with their original embedded tokens verse one with a conditioned conditioned embedded tokens pre-trained on the ImageNet-1k dataset. We report the classification top-1\% accuracy. In each case we see the transformers employing conditioned embedded tokens (ours) outperforms the original ones.}
    \label{tab:vits}
\end{table*}

\begin{remark}\label{rmk:non_lin_cond_assump}
We note that in the case of self-attention with a softmax activation \cref{thm:complexity_lower} requires an assumption on the condition number of the non-linear self-attention probability matrix term $\mathbf{softmax}((X+C)W_QW_K^T(X^T+C^T))$. Although this may be seen as a limitation of the theory we will see in \cref{sec:exps} that this framework still provides a useful methodology to obtain good performance on a variety of transformer applications.
\end{remark}

\begin{remark}\label{rmk:other_layers}
\cref{thm:complexity_lower} demonstrates that conditioning embedded tokens can improve the condition number of the self-attention matrix in the first layer. However, it does not address how this affects the condition number in subsequent layers. In \cref{sec:exps}, we will show that this effect propagates through the network, leading to a reduced condition number in deeper self-attention layers.
\end{remark}

\begin{remark}\label{rmk:comparison_to_lit}
Our theory, as presented in \cref{thm:conditioned_tokens} and \cref{thm:complexity_lower}, demonstrates that conditioning embedded tokens reduces the condition number of the self-attention matrix in the first layer. However, the framework does not theoretically establish how this improvement translates to better optimization and, consequently, higher accuracy in applications. Nonetheless, in \cref{sec:exps}, we provide empirical evidence showing that our approach consistently enhances accuracy across various applications, highlighting its practical effectiveness.
\end{remark}


\section{Experiments}\label{sec:exps}

In this section, we evaluate our insight from \cref{sec:theory} on a variety of transformer applications. For a detailed analysis of how we implemented conditioned embedded tokens see Sec. A in the Supp. material.

\subsection{Image Classification}\label{subsec:IC}

In this section, we apply our theoretical insights from  \cref{sec:theory} to vision transformers on an image classification task.  We will train all vision transformers from scratch on the ImageNet-1k dataset. We will be testing on the following vision transformers. In each case we will compare the original architecture with one that uses conditioned embedded tokens.

\textbf{ViT \cite{dosovitskiy2020image}:} ViT is a pioneering architecture that applies transformer-based processing to images by treating them as sequences of non-overlapping patches. In our study, we utilize ViT-Base (ViT-B), which is configured with a patch size of 16, an embedding dimension of 768, 12 attention heads, and 12 layers. The model employs a standard self-attention mechanism to capture dependencies across patches.

\textbf{DeiT \cite{touvron2021training}:} DeiT builds upon ViT but is optimized for data efficiency. Unlike standard ViT, it employs a data-starvation training strategy \cite{touvron2021training}, enabling faster convergence while maintaining strong performance. In our setup, we use DeiT-Base (DeiT-B) with a patch size of 16, an embedding dimension of 768, 12 attention heads, and 12 layers. Like ViT, DeiT utilizes a self-attention mechanism for feature extraction.

\begin{figure}[ht!]
    \centering
    \includegraphics[width=1.15\linewidth]
    {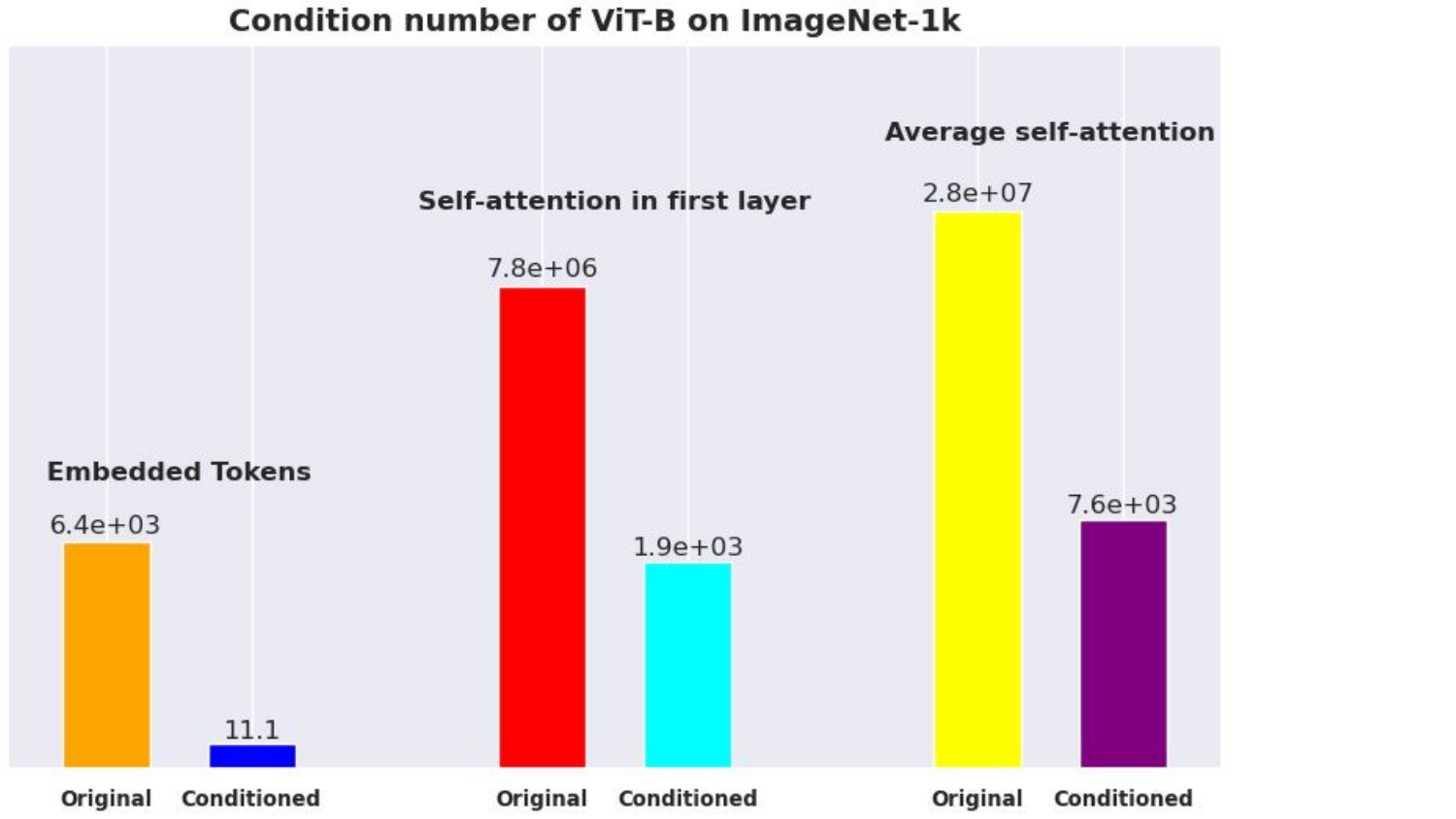}
    \caption{Condition number comparison for ViT-B on ImageNet-1k. The left, middle, and right bars show the condition number of embedded tokens, first-layer self-attention (averaged across heads), and self-attention across all layers, respectively, averaged over training epochs. Our model (conditioned) consistently achieves a significantly lower condition number than the original.}
    \label{fig:vit_b_condition}
\end{figure}

\textbf{Swin Transformer \cite{liu2021swin}:} Swin Transformer introduces a hierarchical architecture and a novel shifted window-based self-attention mechanism, significantly improving efficiency and effectiveness in vision tasks. In our experiments, we adopt the Swin-Base (Swin-B) variant, which features 128 channels in the hidden layers of the first stage. The attention operates over non-overlapping windows of size $M = 7$, while the query dimension per head is set to $d = 32$. The network follows a hierarchical structure with layers arranged as $\{2, 2, 18, 2\}$ across different stages.

\textbf{XCiT \cite{ali2021xcit}:} XCiT diverges from standard ViT architectures by incorporating two key innovations. First, it integrates Local Patch Interaction within each block, using depth-wise $3 \times 3$ convolutions, Batch Normalization, GELU activation, and an additional depth-wise convolution to enhance local feature representation. Second, it employs Cross-Covariance Attention, where attention maps are derived from the cross-covariance matrix of the key and query projections. For our experiments, we utilize the XCiT-M24 (XCiT-M) model with a patch size of 16, an embedding dimension of 512, 8 attention heads, and 24 layers.

\begin{table*}[!ht]

    \centering

    \setlength{\tabcolsep}{14pt}
    \begin{tabular}{c|c c c |c c c}
        \midrule
         Model & $AP^b$ & $AP^b_{50}$ & $AP^b_{75}$ & $AP^m$ & $AP^m_{50}$ & $AP^m_{75}$ \\
        \midrule
        XCiT-S + original embedded tokens & 44.9 & 66.1 & 48.9 & 40.1 & 63.1 & 42.8 \\
        \midrule
        XCiT-S + conditioned embedded tokens & \textbf{45.7} & \textbf{66.6} & \textbf{49.7} & \textbf{40.4} & \textbf{63.5} & \textbf{43.4} \\
        \midrule
        XCiT-M + original embedded tokens & 45.7 & 66.8 & 49.6 & 40.8 & 63.6 & 43.3 \\
        \midrule
        XCiT-M + conditioned embedded tokens & \textbf{46.2} & \textbf{67.4} & \textbf{49.8} & \textbf{41.4} & \textbf{63.8} & \textbf{44.2} \\
        \midrule
    \end{tabular}
\caption{Performance evaluation of object detection and instance segmentation on the COCO mini-val set. The backbone networks are pretrained on ImageNet-1k and integrated into the Mask R-CNN framework. The reported metrics include \( AP^b \) (Average Precision for bounding box predictions), \( AP^b_{50/75} \) (Average Precision at IoU thresholds of 0.50 and 0.75 for bounding boxes), \( AP^m \) (Average Precision for mask predictions), and \( AP^m_{50/75} \) (Average Precision at IoU thresholds of 0.50 and 0.75 for mask predictions). In each case we see the XCiT architecture employing conditioned embedded tokens (ours) out performs the original architecture.}
    \label{tab:transferlearning}
\end{table*}

\textbf{DaViT \cite{ding2022davit}:} DaViT extends traditional vision transformers by incorporating both spatial and channel attention mechanisms, enhancing feature extraction across multiple dimensions. While conventional transformers rely solely on spatial self-attention, DaViT introduces channel attention to capture dependencies across feature channels, complementing spatial interactions. This hybrid approach allows DaViT to effectively model both local and global relationships. In our experiments, we use the DaViT-Base (DaViT-B) model, configured with a patch size of 4, embedding dimensions of $(128, 256, 512, 1024)$, attention heads of $(4, 8, 16, 32)$, and a layer distribution of $(1, 1, 9, 1)$.


\paragraph{Results:} \Cref{tab:vits} presents a comparison of the top-1\% test accuracy of five vision transformers trained from scratch on the ImageNet-1k dataset. Each model is evaluated using both the original embedded tokens and our conditioned embedded tokens, following the approach outlined in \cref{subsec:cond_emb_token}. While ViT-Base \cite{dosovitskiy2020image} and DeiT-Base \cite{touvron2021training} utilize standard self-attention mechanisms, Swin-Base \cite{liu2021swin}, XCiT-Medium \cite{ali2021xcit}, and DaViT-Base \cite{ding2022davit} incorporate more advanced attention mechanisms. We include these latter models to illustrate that, although our theoretical analysis focuses on self-attention, conditioned embedded tokens can be easily integrated into modern transformers with more sophisticated architectures. As shown in the table, models using conditioned embedded tokens consistently achieve higher accuracy. Full implementation details and hyperparameter settings can be found in Sec. B.2 of the Supp. material.

\textbf{Analysis:} \Cref{fig:vit_b_condition} illustrates the condition numbers of the embedded tokens, the first-layer self-attention matrix (averaged across all heads), and the overall self-attention matrix (averaged across all layers) for the ViT-B architecture. We compare these metrics for models using the original embedded tokens and our conditioned embedded tokens. These values were computed at each training epoch and then averaged over all epochs to generate the plots. The figure clearly shows that the ViT-B architecture with conditioned embedded tokens maintains on average a significantly lower condition number throughout training.

\textbf{Implementation and hardware:} For full details on the implementation of each vision transformer, the training hyperparameters and the hardware used please see Sec. B.2 in the Supp. material.

\subsection{Object Detection and Image Segmentation}\label{subsec:OD}

In this section, we extend our approach from \cref{subsec:cond_emb_token} to object detection and instance segmentation tasks. To evaluate its effectiveness in fine-tuning scenarios, we adapt a pretrained XCiT model, originally trained on ImageNet-1k, for these tasks. Our experiments leverage the COCO 2017 dataset \cite{lin2014microsoft}, which consists of 118K training images and 5K validation images spanning 80 object categories. For model architecture, we employ XCiT as the backbone of the Mask R-CNN framework \citep{he2017mask}, incorporating a Feature Pyramid Network (FPN) to enhance multi-scale feature representation. To integrate XCiT with FPN, we modify its inherently columnar structure by extracting features from multiple layers. We do this by using an XCiT-small that has 12 layers (XCiT-S) and a XCiT-medium (XCiT-M) that has 24 layers. These features, originally computed at a fixed stride of 16, are adjusted to strides of \([4, 8, 16, 32]\) to align with the FPN hierarchy. Max pooling is used for downsampling, while a single transposed convolution layer facilitates upsampling. The model is then fine-tuned for 36 epochs using the AdamW optimizer, with a learning rate of \(10^{-4}\), a weight decay of \(0.05\), and a batch size of 16. This training strategy effectively demonstrates how XCiT's learned representations can be adapted for downstream tasks.

\textbf{Results:} We pretrained a total of four models on the ImageNet-1k dataset, namely an XCiT-S and XCiT-M both using the original embedded tokens following the methodology from \cref{subsec:cond_emb_token}.
The results of the experiment are shown in \cref{tab:transferlearning}. As can be seen from the table in both cases the transformer using conditioned embedded tokens out performs the original.

\textbf{Implementation and hardware:} For full details on the implementation, the training hyperparameters and the hardware used see Sec. B.3 in the Supp. material.


\subsection{Language Models}\label{subsec:LM}

\begin{table*}[h]
    \centering
    \begin{tabular}{l|cccccccc|c}
        \toprule
        & MNLI & SST-2 & STSB & RTE & QNLI & QQP & MRPC & CoLA & GLUE \\
        \midrule
        Crammed BERT (original) & 83.8 & 92.3 & 86.3 & 55.1 & 90.1 & 87.3 & 85.0 & 48.9 & 78.6 \\
         \midrule
        Crammed BERT (ours) & \textbf{84.2} & \textbf{92.5} & \textbf{86.5}  & \textbf{55.6}  & \textbf{91.1}  & \textbf{87.4}  & \textbf{86.3}  & \textbf{53.7}  & \textbf{79.7}  \\
        \bottomrule
    \end{tabular}
    \caption{Evaluation of a pre-trained Crammed BERT on the GLUE benchmark with the original embedded tokens (original) and our conditioned embedded tokens (ours). As can be seen from the table our methodology out performs the original in every task.}
    \label{tab:bert_glue_results}
\end{table*}

In this section, we evaluate our insights from \cref{subsec:cond_emb_token} on two different language models.

\paragraph{Crammed BERT:}
In the first experiment we applied our insights from \cref{subsec:cond_emb_token} to a Crammed BERT language model \cite{geiping2023cramming}, trained entirely from scratch using masked language modeling with conditioned embedded tokens. The model consists of 110 million parameters, with 12 transformer layers and 12 attention heads per layer. We train two versions from scratch: the original Crammed BERT \cite{geiping2023cramming} and a variant incorporating conditioned embedded tokens using the insight from \cref{subsec:cond_emb_token}. Both models are trained on The Pile dataset \cite{gao2021pile}, a large-scale corpus designed for language model training following the pretraining regime in \cite{geiping2023cramming}.
After pretraining, we evaluate the performance on the GLUE benchmark \cite{wang2018glue} following the evaluation methodology outlined in \cite{geiping2023cramming}.

\textbf{Results:} The evaluation results are shown in \cref{tab:bert_glue_results} with each number representing the accuracy. We see from the table that in each task the model employing conditioned embedded tokens outperforms the original Crammed Bert model on every task within the GLUE benchmark.

\paragraph{GPT-2:}
We conducted a second experiment training a GPT-2 model with a decoder-only transformer architecture, consisting of 12 layers, 12 attention heads per layer, and an embedding dimension of 512. The model was trained using masked self-attention to predict the next token, leveraging the TinyStories dataset \cite{eldan2023tinystories} for language generation. Two versions of the model were trained from scratch: one using the original embedded tokens as in \cite{eldan2023tinystories} and another incorporating the conditioned embedded tokens from \cref{subsec:cond_emb_token}. Both models were evaluated using a validation loss.


\begin{table}[h]
    \centering
    \begin{tabular}{|c|c|}
        \hline
         & Val loss \\ 
        \hline 
        GPT-2 original & 2.41 \\  
        \hline
        GPT-2 conditioned (ours) & \textbf{2.36} \\  
        \hline
    \end{tabular}
    \caption{GPT-2 model trained on the TinyStories dataset. We compare two models, the original model and one using conditioned embedded tokens (conditioned). We report the validation loss, with the conditioned model achieving lower loss and thus better performance.}
    \label{tab:gpt}
\end{table}

\begin{figure}[ht!]
    \centering
    \includegraphics[width=1.15\linewidth]
    {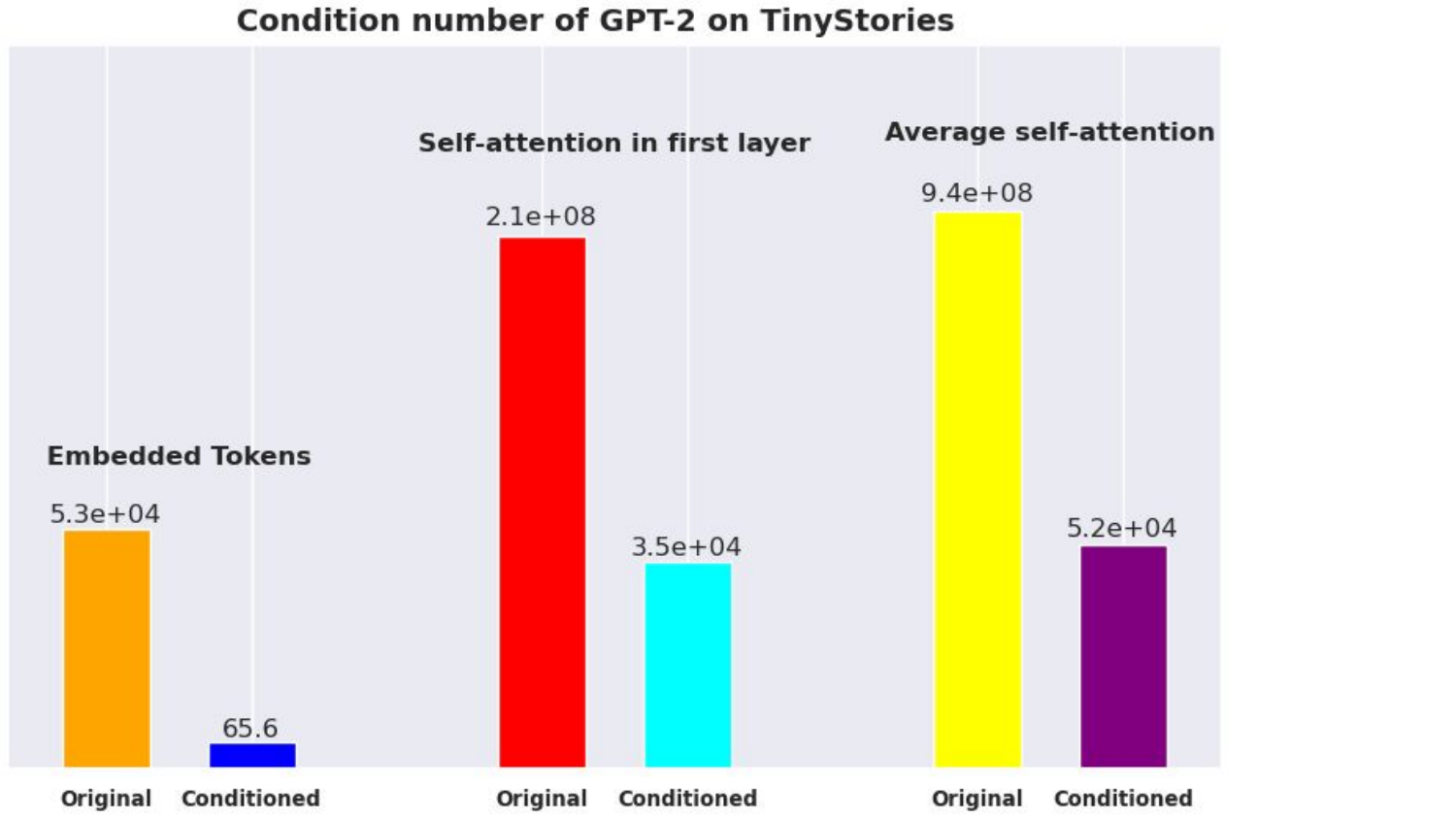}
    \caption{Condition number comparison for GPT-2 training on TinyStories. The left, middle, and right bars show the condition number of embedded tokens, first-layer self-attention (averaged across heads), and self-attention across all layers, respectively, averaged over all epochs. Our model (conditioned) consistently achieves a significantly lower condition number than the original.}
    \label{fig:gpt2}
\end{figure}

\textbf{Results and Analysis:} \cref{tab:gpt} presents the results of our experiment, demonstrating that the model trained with conditioned embedded tokens consistently achieves lower validation loss when compared to the original GPT-2 model. These results indicate that conditioning the embedded tokens improves model performance, leading to more accurate language generation. \cref{fig:gpt2} plots the condition numbers of the embedded tokens, self-attention in first layer averaged over all heads and self-attention averaged over all layers. Each of these are evaluated over each training epoch and then averaged over all epochs.

\paragraph{Implementation and hardware:} For full details on the implementation, the training hyperparameters and the hardware used please see Sec. B.4 in the Supp. material.

\begin{table*}[!ht]
  
    \centering
    \setlength{\tabcolsep}{12pt}
    \begin{tabular}{c|c c c c c }
        \midrule
         Model & ListOps & Text & Retrieval & Image & Pathfinder \\
        \midrule
         Nystr\"omformer (original) & 37.1 & 63.8 & 79.8 & 39.9 & 72.9 \\
        \midrule
        Nystr\"omformer conditioned (ours) & \textbf{37.9} & \textbf{64.9} & \textbf{80.9} & \textbf{40.1} & \textbf{73.3} \\
        \midrule
    \end{tabular}
\caption{Comparison of a Nystr\"omformer using its original embedded tokens (original) with a Nystr\"omformer using conditioned embedded tokens (ours) on the LRA benchmark. We report the evaluation accuracy (\%). As can be seen from the table our methodology yields a higher accuracy on each task.}
    \label{tab:nystrom}
\end{table*}

\subsection{Nystr\"omformer for Long Range Sequences}




We applied our methodology to the Nystr\"omformer \cite{xiong2021nystromformer}, a transformer architecture designed for efficient long-range dependency modeling on the Long-Range Arena (LRA) benchmark \cite{tay2020long}. We trained two Nystr\"omformer models: one utilizing conditioned embedded tokens and the original Nystr\"omformer for comparison. The Long-Range Arena (LRA) benchmark consists of five distinct tasks:  
\begin{enumerate}
    \item \textbf{ListOps}: A synthetic task that tests hierarchical reasoning by evaluating nested arithmetic operations.
    \item \textbf{Text Classification}: Character-level text classification, requiring models to process long sequences of text.
    \item \textbf{Retrieval}: A document retrieval task that measures the model’s ability to identify relevant patterns within long textual inputs.
    \item \textbf{Image Classification}: Modeling image data as a sequence of flattened pixel values to assess a model’s capability in visual recognition.
    \item \textbf{Pathfinder}: A visual reasoning challenge that requires detecting connected paths in an image, emphasizing spatial relationship modeling.
\end{enumerate}

\begin{figure}[ht!]
    \centering
    \includegraphics[width=1.15\linewidth]
    {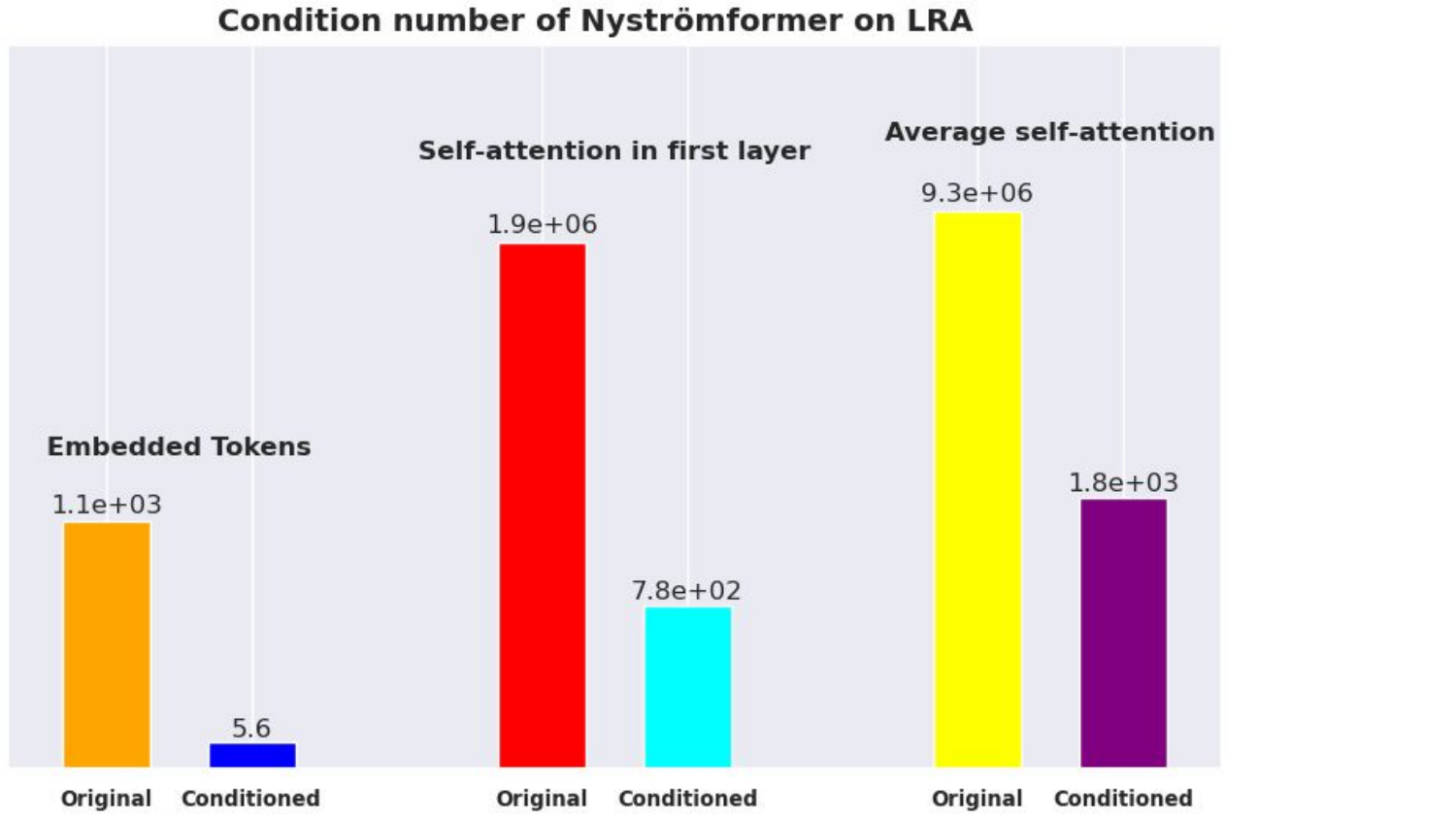}
    \caption{Condition number comparison for Nystr\"omformer on the LRA benchmark. The left, middle, and right bars show the condition number of embedded tokens, first-layer self-attention (averaged across heads), and self-attention across all layers, averaged over training epochs and tasks. Our model (conditioned) consistently achieves a significantly lower condition number than the original.}
    \label{fig:nyst}
\end{figure}

\paragraph{Results and Analysis:} \cref{tab:nystrom} shows the results of the experiment. As can be seen from the table the Nystr\"omformer that employed conditioned embedded tokens outperformed the original Nystr\"omformer from \cite{xiong2021nystromformer} in every task in the LRA benchmark.
We computed the condition numbers of the original Nystr\"omformer and one employing conditioned embedded tokens throughout training for the embedded tokens, the self-attention within the first layer averaged over all heads and the self-attention averaged over all layers. We then took the average for each task over all training epochs and averaged over the five tasks within the LRA benchmark. The plots are shown in \cref{fig:nyst}. As can be seen from the figure, the Nystr\"omformer employing conditioned embedded tokens has lower condition number in all cases. 

\textbf{Implementation and hardware:} For full details on the implementation, the training hyperparameters and the hardware used please see Sec. B.5 in the Supp. material.

\section{Limitations}
Our work explored the conditioning of transformer attention blocks through embedded tokens. While we showed that conditioned embedded tokens reduce the first-layer self-attention condition number, their impact on optimization remains theoretically unproven. Prior studies \cite{liu2022loss, agarwal2021deep, jacot2018neural} have linked NTK conditioning to optimization in feedforward networks, raising an important question: how does conditioning embedded tokens affect the NTK in transformers? Investigating this could offer deeper insights into how embedded token structure influences transformer optimization. 


\section{Conclusion}
In this paper, we introduced a framework for analyzing the conditioning of the self-attention matrix in the first layer of a transformer, highlighting its dependence on the embedded tokens. Our analysis demonstrated that by adding a carefully designed correction term—which we refer to as conditioned embedded tokens—we could significantly reduce the condition number of self-attention in the first layer, leading to a better-conditioned attention mechanism. We empirically validated the effectiveness of conditioned embedded tokens across a diverse set of tasks, including image classification, object detection, instance segmentation, language modeling, and long-range sequence modeling. In all cases, our approach proved to be a simple, drop-in replacement for existing embedding methods, consistently improving model performance across these applications.


\clearpage
{
    \small
    \bibliographystyle{ieeenat_fullname}
    \bibliography{main}
}

\end{document}